\documentclass{llncs}

\usepackage{cite}
\usepackage[pdftex]{graphicx}
\usepackage[export]{adjustbox}

\usepackage{hyperref}

\usepackage{mypackage}

\usepackage{lineno}

\newcommand{\anonymize}[1]\

\begin{document}

\title{Recognizing Surgical Activities with\\Recurrent Neural Networks}

\author{Robert DiPietro\inst{1} \and
	    Colin Lea\inst{1} \and
	    Anand Malpani\inst{1} \and
	    Narges Ahmidi\inst{1} \and\\
	    S. Swaroop Vedula\inst{1} \and
	    Gyusung I. Lee\inst{2} \and
	    Mija R. Lee\inst{2} \and
	    Gregory D. Hager\inst{1}}
	    

\institute{Department of Computer Science, Johns Hopkins University, Baltimore, MD, USA \and
           Department of Surgery, Johns Hopkins University, Baltimore, MD, USA}

%

\maketitle


\begin{abstract}
We apply recurrent neural networks to the task of recognizing surgical activities from robot kinematics. Prior work in this area focuses on recognizing short, low-level activities, or \emph{gestures}, and has been based on variants of hidden Markov models and conditional random fields. In contrast, we work on recognizing both gestures and longer, higher-level activites, or \emph{maneuvers}, and we model the mapping from kinematics to gestures/maneuvers with recurrent neural networks. To our knowledge, we are the first to apply recurrent neural networks to this task. Using a single model and a single set of hyperparameters, we match state-of-the-art performance for gesture recognition and advance state-of-the-art performance for maneuver recognition, in terms of both accuracy and edit distance. Code is available at \url{https://github.com/rdipietro/miccai-2016-surgical-activity-rec}.
\end{abstract}

\setcounter{footnote}{0}

\section{Introduction} \label{sec:introduction}

\begin{figure}[t]
	\centering
	\includegraphics[scale=0.65]{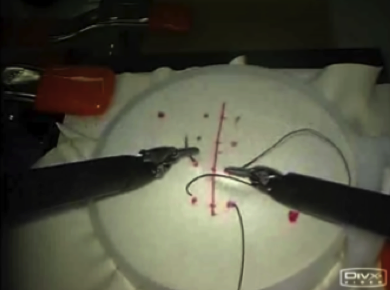} \qquad
	\includegraphics[scale=0.65]{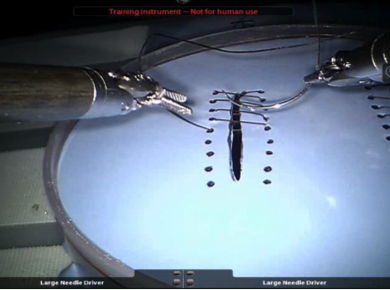}
	\caption{Example images from the JIGSAWS and MISTIC datasets.}
	\label{fig:jigsaws}
\end{figure}

Automated surgical-activity recognition is a valuable precursor for higher-level goals such as objective surgical-skill assessment and for providing targeted feedback to trainees. Previous research on automated surgical-activity recognition has focused on gestures within a surgical task \cite{lea2016learning}, \cite{tao2013surgical}, \cite{lea2015improved}, \cite{sefati2015learning}. Gestures are atomic segments of activity that typically last for a few seconds, such as grasping a needle. In contrast, maneuvers are composed of a sequence of gestures and represent higher-level segments of activity, such as tying a knot. We believe that targeted feedback for maneuvers is meaningful and consistent with the subjective feedback that faculty surgeons currently provide to trainees.

Here we focus on jointly segmenting and classifying surgical activities. Other work in this area has focused on variants of hidden Markov models (HMMs) and conditional random fields (CRFs) \cite{lea2016learning}, \cite{tao2013surgical}, \cite{lea2015improved}, \cite{sefati2015learning}. HMM and CRF based methods often define unary (label-input) and pairwise (label-label) energy terms, and during inference find a global label configuration that minimizes overall energy. Here we put emphasis on the unary terms and note that defining unaries that are both general and meaningful is a difficult task. For example, of the works above, the unaries of \cite{lea2016learning} are perhaps most general: they are computed using \emph{learned} convolutional filters. However, we note that even these unaries depend only on inputs from fairly local neighborhoods in time.

In this work, we use recurrent neural networks (RNNs), and in particular long short-term memory (LSTM), to map kinematics to labels. Rather than operating only on local neighborhoods in time, LSTM maintains a memory cell and \emph{learns} when to write to memory, when to reset memory, and when to read from memory, forming unaries that in principle depend on \emph{all} inputs. In fact, we will rely \emph{only} on these unary terms, or in other words assume that labels are independent given the sequence of kinematics. Despite this, we will see that predicted labels are smooth over time with no post-processing. Further, using a single model and a single set of hyperparameters, we match state-of-the-art performance for gesture recognition and improve over state-of-the-art performance for maneuver recognition, in terms of both accuracy and edit distance.

\section{Methods}\label{sec:methods}

The goal of this work is to use $n_x$ kinematic signals over time to label every time step with one of $n_y$ surgical activities. An individual sequence of length $T$ is composed of kinematic inputs $\{x_t\}$, with each $x_t \in \mathbb{R}^{n_x}$, and a collection of one-hot encoded activity labels $\{y_t\}$, with each $y_t \in \{0, 1\}^{n_y}$. (For example, if we have classes 1, 2, and 3, then the one-hot encoding of label 2 is $(0, 1, 0)^T$.) We aim to learn a mapping from $\{x_t\}$ to $\{y_t\}$ in a supervised fashion that generalizes to users that were absent from the training set. In this work, we use recurrent neural networks to discriminatively model $p(y_t | x_1, x_2, \ldots, x_t)$ for all $t$ when operating online and $p(y_t | x_1, x_2, \ldots, x_T)$ for all $t$ when operating offline.

\subsection{Recurrent Neural Networks}

Though not yet as ubiquitous as their feedforward counterparts, RNNs have been applied successfully to many diverse sequence-modeling tasks, from text-to-handwriting generation \cite{graves2013generating} to machine translation \cite{sutskever2014sequence}.
\begin{figure}[t]
	\centering
	\begin{subfigure}[b]{0.55\textwidth}
		\centering
		\includegraphics[scale=0.7]{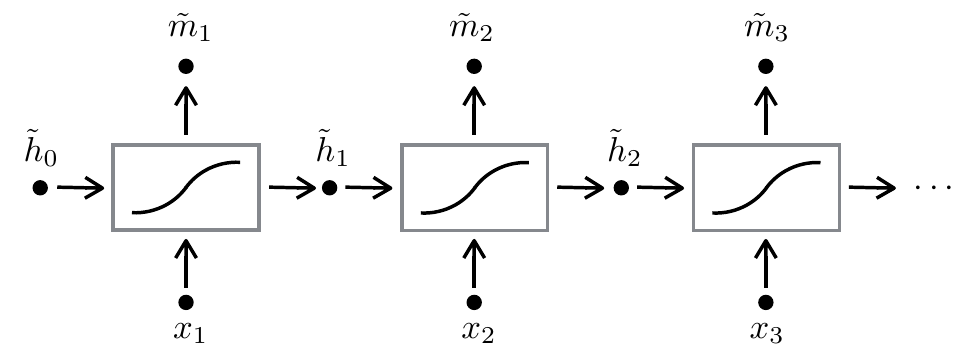}
		\caption{A recurrent neural network.}
		\label{fig:rnn}
	\end{subfigure}
	\hfill
	\begin{subfigure}[b]{0.44\textwidth}
		\centering
		\includegraphics[scale=0.7]{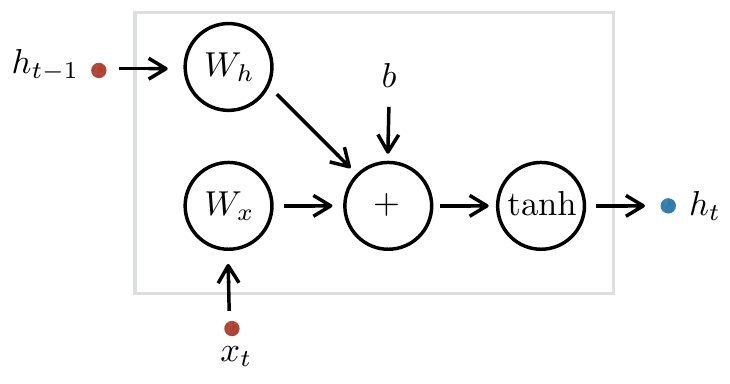}
		\caption{A vanilla RNN block.}
		\label{fig:vanillablock}
	\end{subfigure}
	\caption{The generic RNN on the left has a nonlinear block that is not yet specified and is therefore capable of representing many different models. The nonlinear block on the right yields a vanilla RNN.}
	\label{fig:rnnvanillablock}
\end{figure}

A generic RNN is shown in Figure \ref{fig:rnn}. An RNN maintains a hidden state $\tilde{h}_t$, and at each time step $t$, the nonlinear block uses the previous hidden state $\tilde{h}_{t-1}$ and the current input $x_t$ to produce a new hidden state $\tilde{h}_t$ and an output $\tilde{m}_t$.

If we use the nonlinear block shown in Figure \ref{fig:vanillablock}, we end up with a specific and simple model: a vanilla RNN with one hidden layer. The recursive equation for a vanilla RNN, which can be read off precisely from Figure \ref{fig:vanillablock}, is
\begin{equation}
	h_t = \tanh(W_x x_t + W_h h_{t-1} + b) \label{eq:vanillarnn}
\end{equation}
Here, $W_x$, $W_h$, and $b$ are free parameters that are shared over time. For the vanilla RNN, we have $\tilde{m}_t = \tilde{h}_t = h_t$. The height of $h_t$ is a hyperparameter and is referred to as the number of hidden units.

In the case of multiclass classification, we use a linear layer to transform $\tilde{m}_t$ to appropriate size $n_y$ and apply a softmax to obtain a vector of class probabilities:
\begin{gather}
	\hat{y}_t = \mbox{softmax}(W_{ym} \tilde{m}_t + b_y) \label{eq:logits} \\
	p(y_{tk} = 1~|~x_1, x_2, \ldots, x_t) = \hat{y}_{tk} \label{eq:prob}
\end{gather}
where $\mbox{softmax}(x) = \exp(x) \label{eq:softmax} / \sum_i \exp(x_i)$.

RNNs traditionally propagate information forward in time, forming predictions using only past and present inputs. Bidirectional RNNs \cite{schuster1997bidirectional} can improve performance when operating offline by using future inputs as well. This essentially consists of running one RNN in the forward direction and one RNN in the backward direction, concatenating hidden states, and computing outputs jointly.

\subsection{Long Short-Term Memory}

Vanilla RNNs are very difficult to train because of what is known as the \emph{vanishing gradient problem} \cite{bengio1994learning}. LSTM \cite{hochreiter1997long} was specifically designed to overcome this problem and has since become one of the most widely-used RNN architectures. The recursive equations for the LSTM block used in this work are
\begin{align}
	\tilde{x}_t &= \tanh(W_{\tilde{x}x} x_t + W_{\tilde{x}m} m_{t-1} + b_{\tilde{x}}) \\
	i_t &= \sigma(W_{ix} x_t + W_{im} m_{t-1} + W_{ic} c_{t-1} + b_i) \\
	f_t &= \sigma(W_{fx} x_t + W_{fm} m_{t-1} + W_{fc} c_{t-1} + b_f) \\
	c_t &= i_t \odot \tilde{x}_t + f_t \odot c_{t-1} \\
	o_t &= \sigma(W_{ox} x_t + W_{om} m_{t-1} + W_{oc} c_t + b_o) \\
	m_t &= o_t \odot \tanh(c_t)
\end{align}
where $\odot$ represents element-wise multiplication and $\sigma(x) = 1 / (1+\exp(-x))$. All matrices $W$ and all biases $b$ are free parameters that are shared across time.

LSTM maintains a memory over time and \emph{learns} when to write to memory, when to reset memory, and when to read from memory \cite{graves2012supervised}. In the context of the generic RNN, $\tilde{m}_t = m_t$, and $\tilde{h}_t$ is the concatenation of $c_t$ and $m_t$. $c_t$ is the \emph{memory cell} and is updated at each time step to be a linear combination of $\tilde{x}_t$ and $c_{t-1}$, with proportions governed by the \emph{input gate} $i_t$ and the \emph{forget gate} $f_t$. $m_t$, the output, is a nonlinear version of $c_t$ that is filtered by the output gate $o_t$. Note that all elements of the gates $i_t$, $f_t$, and $o_t$ lie between 0 and 1.

This version of LSTM, unlike the original, has forget gates and \emph{peephole connections}, which let the input, forget, and output gates depend on the memory cell. Forget gates are a standard part of modern LSTM \cite{greff2015lstm}, and we include peephole connections because they have been found to improve performance when precise timing is required \cite{gers2000recurrent}. All weight matrices are full except the peephole matrices $W_{ic}$, $W_{fc}$, and $W_{oc}$, which by convention are restricted to be diagonal.

\subsubsection{Loss}

Because we assume every $y_t$ is independent of all other $y_{t'}$ given $x_1, \ldots, x_t$, maximizing the log likelihood of our data is equivalent to minimizing the overall cross entropy between the true labels $\{y_t\}$ and the predicted labels $\{\hat{y}_t\}$. The global loss for an individual sequence is therefore
\begin{equation*}
l_{\mbox{seq}}(\{y_t\}, \{\hat{y}_t\}) = \sum_t l_t(y_t, \hat{y}_t) \qquad \mbox{with} \qquad
l_t(y_t, \hat{y}_t) = -\sum_k y_{tk} \log \hat{y}_{tk}
\end{equation*}

\subsubsection{Training}

All experiments in this paper use standard stochastic gradient descent to minimize loss. Although the loss is non-convex, it has repeatedly been observed empirically that ending up in a poor local optimum is unlikely. Gradients can be obtained efficiently using backpropagation \cite{rumelhart1988learning}. In practice, one can build a computation graph out of fundamental operations, each with known local gradients, and then apply the chain rule to compute overall gradients with respect to all free parameters. Frameworks such as Theano and Google TensorFlow let the user specify these computation graphs symbolically and alleviate the user from computing overall gradients manually.

Once gradients are obtained for a particular free parameter $p$, we take a small step in the direction opposite to that of the gradient: with $\eta$ being the learning rate,
\begin{equation*}
p' = p - \eta \frac{\partial l_{\mbox{seq}}}{\partial p} \qquad \mbox{with} \qquad \frac{\partial l_{\mbox{seq}}}{\partial p} = \sum_t \frac{\partial l_t}{\partial p}
\end{equation*}

\section{Experiments}

\subsection{Datasets}

The JHU-ISI Gesture and Skill Assessment Working Set (JIGSAWS) \cite{gao2014} is a public benchmark surgical activity dataset recorded using the \emph{da Vinci}. JIGSAWS contains synchronized video and kinematic data from a standard 4-throw suturing task performed by eight subjects with varying skill levels. All subjects performed about 5 trials, resulting in a total of 39 trials. We use the same measurements and activity labels as the current state-of-the-art method \cite{lea2016learning}. Measurements are position ($x$, $y$, $z$), velocity ($v_x$, $v_y$, $v_z$), and gripper angle ($\theta$) for each of the left and right slave manipulators, and the surgical activity at each time step is one of ten different gestures.

The Minimally Invasive Surgical Training and Innovation Center - Science of Learning (MISTIC-SL) dataset, also recorded using the \emph{da Vinci}, includes 49 right-handed trials performed by 15 surgeons with varying skill levels. We follow \cite{gao2016unsupervised} and use a subset of 39 right-handed trials for all experiments. All trials consist of a suture throw followed by a surgeon's knot, eight more suture throws, and another surgeon's knot. We used the same kinematic measurements as for JIGSAWS, and the surgical activity at each time step is one of 4 maneuvers: suture throw (ST), knot tying (KT), grasp pull run suture (GPRS), and inter-maneuver segment (IMS). It is not possible for us to release this dataset at this time, though we hope we will be able to release it in the future.

\subsection{Experimental Setup}

JIGSAWS has a standardized leave-one-user-out evaluation setup: for the $i$-th run, train using all users except $i$ and test on user $i$. All results in this paper are averaged over the 8 runs, one per user. We follow the same strategy for MISTIC-SL, averaging over 11 runs, one for each user that does not appear in the validation set, as explained below.

We include accuracy and edit distance (Levenshtein distance) as performance metrics. Accuracy is the percentage of correctly-classified frames, measuring performance without taking temporal consistency into account. In contrast, edit distance is the number of operations needed to transform predicted segment-level labels into ground-truth segment-level labels, here normalized for each dataset using the maximum number (over all sequences) of segment-level labels.

\subsection{Hyperparameter Selection and Training}

Here we include the most relevant details regarding hyperparameter selection and training; other details are fully specified in code, available at\\ \url{https://github.com/rdipietro/miccai-2016-surgical-activity-rec}.

For each run we train for a total of approximately 80 epochs, maintaining a learning rate of 1.0 for the first 40 epochs and then halving the learning rate every 5 epochs for the rest of training. Using a small batch size is important; we found that otherwise the lack of stochasticity let us converge to bad local optima. We use a batch size of 5 sequences for all experiments.

Because JIGSAWS has a fixed leave-one-user-out test setup, with all users appearing in the test set exactly once, it is not possible to use JIGSAWS for hyperparameter selection without inadvertently training on the test set. We therefore choose all hyperparameters using a small MISTIC-SL validation set consisting of 4 users (those with only one trial each), and we use the resulting hyperparameters for both JIGSAWS experiments and MISTIC-SL experiments. We performed a grid search over the number of RNN hidden layers (1 or 2), the number of hidden units per layer (64, 128, 256, 512, or 1024), and whether dropout \cite{zaremba2014recurrent} is used (with $p = 0.5$). 1 hidden layer of 1024 units, with dropout, resulted in the lowest edit distance and simultaneously yielded high accuracy. These hyperparameters were used for all experiments.

Using a modern GPU, training takes about 1 hour for any particular JIGSAWS run and about 10 hours for any particular MISTIC-SL run (MISTIC-SL sequences are approximately 10x longer than JIGSAWS sequences). We note, however, that RNN inference is fast, with a running time that scales linearly with sequence length. At test time, it took the bidirectional RNN approximately 1 second of compute time per minute of sequence (300 time steps).

\begin{table*}[t]
\centering
\caption{Quantitative results and comparisons to prior work.}
\label{tab:results}
\renewcommand{\arraystretch}{1.2}
\begin{tabularx}{\textwidth}{@{}lcccccc@{}}
\toprule
                                 & \phantom{x} & \multicolumn{2}{c}{\phantom{xxxxxxxxx}JIGSAWS\phantom{xxxxxxxx}}  & \phantom{xx} & \multicolumn{2}{c}{\phantom{xxxxxxxx}MISTIC-SL\phantom{xxxxxxxx}}  \\
                                                 \cmidrule{3-4}                                                                     \cmidrule{6-7}
                                 &              & Accuracy (\%)            & Edit Dist. (\%)               &              & Accuracy (\%)            & Edit Dist. (\%)          \\[1.5ex]
MsM-CRF \cite{tao2013surgical}   &              & 72.6                     & ---                           &              & ---                      & ---                      \\
SDSDL \cite{sefati2015learning}  &              & 78.7                     & ---                           &              & ---                      & ---                      \\
SC-CRF \cite{lea2015improved}    &              & 80.3                     & ---                           &              & ---                      & ---                      \\
LC-SC-CRF \cite{lea2016learning} &              & 82.5 $\pm$ 5.4           & 14.8 $\pm$ 9.4                &              & 81.7 $\pm$ 6.2           & 29.7 $\pm$ 6.8           \\[1.5ex]
Forward LSTM                     &              & 80.5 $\pm$ 6.2           & 19.8 $\pm$ 8.7                &              & 87.8 $\pm$ 3.7           & 33.9 $\pm$ 13.3          \\
Bidir. LSTM                      &              & \textbf{83.3 $\pm$ 5.7}  & \textbf{14.6 $\pm$ 9.6}       &              & \textbf{89.5 $\pm$ 4.0}  & \textbf{19.5 $\pm$ 5.2}  \\[1.5ex]
\bottomrule
\end{tabularx}
\end{table*}

\subsection{Results}

Table \ref{tab:results} shows results for both JIGSAWS (gesture recognition) and MISTIC-SL (maneuver recognition). A forward LSTM and a bidirectional LSTM are compared to the Markov/semi-Markov conditional random field (MsM-CRF), Shared Discriminative Sparse Dictionary Learning (SDSDL), Skip-Chain CRF (SC-CRF), and Latent-Convolutional Skip-Chain CRF (LC-SC-CRF). We note that the LC-SC-CRF results were computed by the original author, using the same MISTIC-SL validation set for hyperparameter selection.

We include standard deviations where possible, though we note that they largely describe the user-to-user variations in the datasets. (Some users are exceptionally challenging, regardless of the method.) We also carried out statistical-significance testing using a paired-sample permutation test ($p$-value of 0.05). This test suggests that the accuracy and edit-distance differences between the bidirectional LSTM and LC-SC-CRF are insignificant in the case of JIGSAWS but are significant in the case of MISTIC-SL. We also remark that even the forward LSTM is competitive here, despite being the only algorithm that can run online.

\begin{figure}[t]
	\centering
	\begin{subfigure}[b]{1.0\textwidth}
		\centering
		\includegraphics[valign=t]{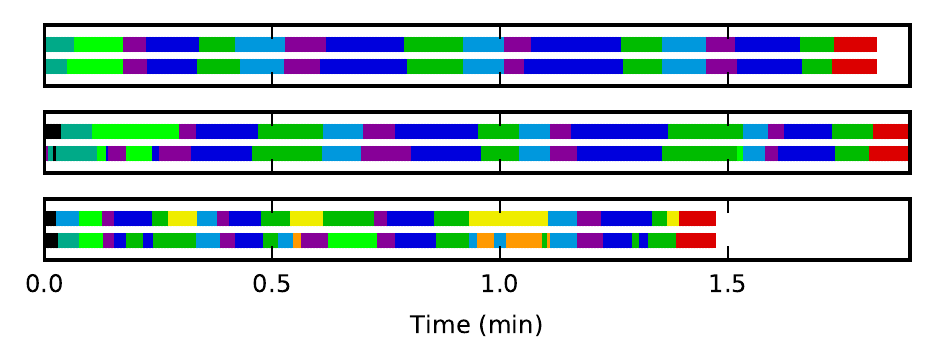}
		\includegraphics[valign=t]{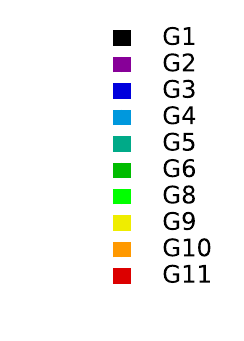}
	\end{subfigure}
	\hfill
	\begin{subfigure}[b]{1.0\textwidth}
		\centering
		\includegraphics[valign=t]{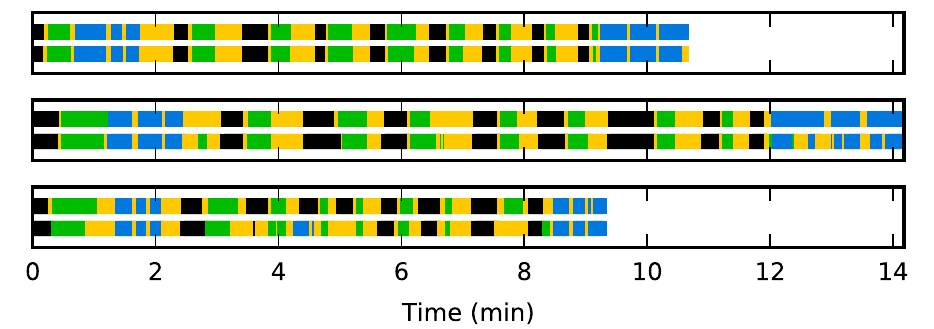}
		\includegraphics[valign=t]{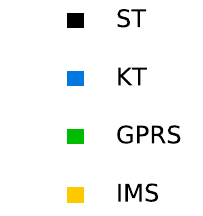}
	\end{subfigure}
	\caption{Qualitative results for JIGSAWS (top) and MISTIC-SL (bottom) using a bidirectional LSTM. For each dataset, we show results from the trials with highest accuracy (top), median accuracy (middle), and lowest accuracy (bottom). In all cases, ground truth is displayed above predictions.}
	\label{fig:results}
\end{figure}

Qualitative results are shown in Figure \ref{fig:results} for the trials with highest, median, and lowest accuracies for each dataset. We note that the predicted label sequences are smooth, despite the fact that we assumed that labels are independent given the sequence of kinematics.

\section{Summary}

In this work we performed joint segmentation and classification of surgical activities from robot kinematics. Unlike prior work, we focused on high-level maneuver prediction in addition to low-level gesture prediction, and we modeled the mapping from inputs to labels with recurrent neural networks instead of with HMM or CRF based methods. Using a single model and a single set of hyperparameters, we matched state-of-the-art performance for JIGSAWS (gesture recognition) and advanced state-of-the-art performance for MISTIC-SL (maneuver recognition), in the latter case increasing accuracy from 81.7\% to 89.5\% and decreasing normalized edit distance from 29.7\% to 19.5\%.

\bibliographystyle{splncs03}
\bibliography{others,los}

\end{document}